\documentclass[runningheads]{llncs}

\usepackage[T1]{fontenc}
\usepackage{graphicx}
\usepackage{booktabs}
\usepackage{amsmath}
\usepackage{amssymb}
\usepackage{url}
\usepackage{hyperref}

\newif\ifblindreview
\newcommand{\osfifty}{\texttt{OS50}}
\newcommand{\aura}{\texttt{AURA}}
\newcommand{\method}{\texttt{OS50+AURA}}

\begin{document}

\title{Asymmetric Paired-Annotation Learning for Multi-Structure ULF Pediatric Brain MRI Segmentation}

% \title{Asymmetric Paired-Annotation Learning for Multi-Structure ULF Pediatric Brain MRI Segmentation}
\titlerunning{Paired-Annotation Learning for ULF MRI}

\ifblindreview
\author{Anonymous Authors}
\authorrunning{Anonymous Authors}
\institute{Anonymous Institution}
\else
\author{
Ha-Hieu Pham\inst{1}\textsuperscript{*} \and
Dang P.M. Cao\inst{1,2}\textsuperscript{*} \and
Minh Hoang Pham\inst{3} \and
Khanh Nguyen Vo Ngoc\inst{4} \and
Thanh-Huy Nguyen\inst{5} \and
Ulas Bagci\inst{6} \and
Huy-Hieu Pham\inst{1,2,7}\textsuperscript{\textdagger}
}

\authorrunning{H.-H. Pham, D.P.M. Cao et al.}

\institute{
VinUni-Illinois Smart Health Center, VinUniversity, Hanoi, Vietnam
\and
College of Engineering \& Computer Science, VinUniversity, Hanoi, Vietnam
\and
Hanoi University of Science and Technology, Hanoi, Vietnam
\and
University of Information Technology, VNU-HCM, Ho Chi Minh City, Vietnam
\and
Carnegie Mellon University, Pittsburgh, PA, USA
\and
Northwestern University, Evanston, IL, USA
\and
Center for Innovations in Health Sciences, VinUniversity, Hanoi, Vietnam
}
\fi

\maketitle

\begingroup
\renewcommand{\thefootnote}{*}
\footnotetext{Ha-Hieu Pham and Dang P. M. Cao contributed equally to this work.}
\renewcommand{\thefootnote}{\textdagger}
\footnotetext{Corresponding author: hieu.ph@vinuni.edu.vn}
\endgroup

\begin{abstract}

Portable ultra-low-field (ULF) MRI can expand access to pediatric neuroimaging, but segmentation at 0.064~T remains challenging because anatomical boundaries are weakly delineated, small structures may be only partially visible, and high-field references can be locally misregistered. The LISA 2026 Challenge provides two non-equivalent annotations reflecting different sources of anatomical evidence: a high-field-derived (HF) mask defining the scored target and a low-field-edited (LF) mask aligned with visible ULF anatomy. In this challenge report, we describe AURA, an nnU-Net-based asymmetric supervision strategy that treats these annotations as distinct observations rather than interchangeable ground truths. AURA anchors training to the HF mask and incorporates the LF mask through a bounded reliability gate based on label disagreement, boundaries, predictive uncertainty, class reliability, and training stage. On a 16-case development split, the HF-supervised baseline, AURA, and their ensemble achieved Dice scores of 0.7984, 0.7950, and 0.7988, respectively, while the ensemble achieved an HD95 of 1.8892 and an ASSD of 0.7855. These results provide a preliminary evaluation of AURA within the LISA 2026 Challenge and motivate further assessment on the hidden test set and external ULF cohorts. Our code and pretrained models are available at 
\url{https://github.com/minhdang050806/A-nnU-Net-based-asymmetric-supervision-strategy}.
\end{abstract}

\section{Introduction}

Portable ultra-low-field (ULF) MRI offers a practical route to pediatric neuroimaging in settings where conventional high-field scanners are costly, inaccessible, or difficult to deploy. Recent work on low-field and portable MRI has emphasized its potential for point-of-care neuroimaging, bedside deployment, and improved access in resource-constrained environments~\cite{arnold2023lowfield,kimberly2023brain,mazurek2021portable,sien2023feasibility}. This accessibility is clinically attractive, but it also introduces challenges for image analysis: at 0.064~T, brain MRI has lower signal-to-noise ratio, weaker tissue contrast, stronger partial-volume effects, and greater sensitivity to motion than high-field MRI~\cite{arnold2023lowfield,kimberly2023brain,billot2023synthseg}. These factors can make multi-structure segmentation difficult, particularly for small deep nuclei, hippocampi, and thin anatomical boundaries whose appearance may be only partially expressed in the low-field image~\cite{lepore2025lisa}. The LISA 2026 segmentation task considers this setting by requiring segmentation of 11 structures from 0.064~T T2-weighted pediatric brain MRI: bilateral hippocampi, lateral ventricles, caudate nuclei, lentiform nuclei, and thalami, together with the corpus callosum~\cite{lisa2026challenge}. Although U-Net-style volumetric segmentation and nnU-Net provide strong baselines for biomedical segmentation~\cite{ronneberger2015unet,cicek20163dunet,isensee2021nnunet}, the paired annotations provided in this task introduce an additional supervision-related consideration.

Each training image is released with two annotations that reflect different observation processes. The high-field-derived (HF) mask is obtained from high-field anatomy registered to the ULF image and defines the official target, but it may be locally affected by registration error. The low-field-edited (LF) mask is intended to reflect structures visible in the ULF acquisition, but it may omit or simplify anatomy that cannot be resolved confidently at low field strength. This setting is related to problems of label fusion, inter-observer variability, and noisy supervision in medical image analysis~\cite{warfield2004staple,karimi2020noisy}, while also introducing an asymmetry between the two annotations in terms of their role in scoring, anatomical visibility, and image alignment. Using only the LF annotation would move training away from the official target, whereas treating the two masks as uniformly reliable may not fully account for their different sources of information. We therefore explore an asymmetric formulation in which the HF mask remains the primary supervision target and the LF mask is incorporated as a conditional auxiliary signal.

Our method, AURA, implements this formulation through a bounded reliability gate. We use the name AURA to denote an asymmetric, uncertainty- and reliability-aware supervision strategy for the paired HF/LF annotation setting. Its intended use case is training with paired annotations while retaining an ULF-only deployment pathway. AURA therefore does not require a high-field image, an HF mask, or an LF mask at test time. The LF contribution is modulated using label disagreement, anatomical boundaries, predictive uncertainty, class-specific reliability, and training stage, with the aim of incorporating potentially complementary LF information while limiting its influence on the primary HF target. The segmentation backbone remains a 3D full-resolution nnU-Net~\cite{isensee2021nnunet}. A hard-voxel baseline is trained on the HF target with soft Dice and TopK cross-entropy, while AURA is initialized from this baseline and fine-tuned using asymmetric paired-label supervision. At inference, both branches receive only the ULF MRI and their probability maps are averaged before argmax prediction and per-class largest connected component filtering.

This report describes the approach developed for the LISA 2026 Challenge and evaluates it within the available development setting. \textbf{First}, we formulate the paired HF/LF annotations as asymmetric observations with different roles in scoring, image alignment, and anatomical visibility. \textbf{Second}, we describe a bounded, reliability-gated supervision strategy that incorporates LF edits while retaining the HF target as the primary supervision source and preserving an ULF-only deployment pathway. \textbf{Third}, we report a reproducible nnU-Net-compatible training, ensembling, and post-processing recipe with evaluation across overlap, surface-distance, and volume-bias metrics, following the recommendation that segmentation validation should combine region-, distance-, and volume-based criteria~\cite{taha2015metrics}. Since the model configuration and ensemble weighting were selected using the available development data, the reported results are intended as a challenge-cycle evaluation rather than evidence of independent generalization. The relatively small performance differences observed on the development setting further motivate cautious interpretation and evaluation on the hidden challenge test set.

\section{Challenge Overview}

\subsection{Task Definition and Released Data}

LISA 2026 Task~2 targets multi-structure segmentation in pediatric ultra-low-field (ULF) brain MRI. Each input is a combined isotropic T2-weighted volume acquired with the 0.064~T Hyperfine Swoop system \cite{lisa2026challenge}, reconstructed from orthogonal ULF acquisitions and linearly registered to a subject- and age-matched high-field (HF) scan. The released HF-derived segmentation defines the official target, while a low-field-edited (LF) mask is provided as an optional annotation aligned with structures visible in the ULF image. The HF images are not released. The task includes 11 foreground labels: bilateral hippocampi, lateral ventricles, caudate nuclei, lentiform nuclei, and thalami, plus the corpus callosum. Background is label 0. We use the 79 labeled cases with the fixed nnU-Net fold-0 split: 63 cases for training and 16 for development validation. Each training case contains both the scored HF target and its paired LF annotation, and both component models are trained on the same 63 cases.

\begin{figure}[!htbp]
\centering
\includegraphics[width=\textwidth]{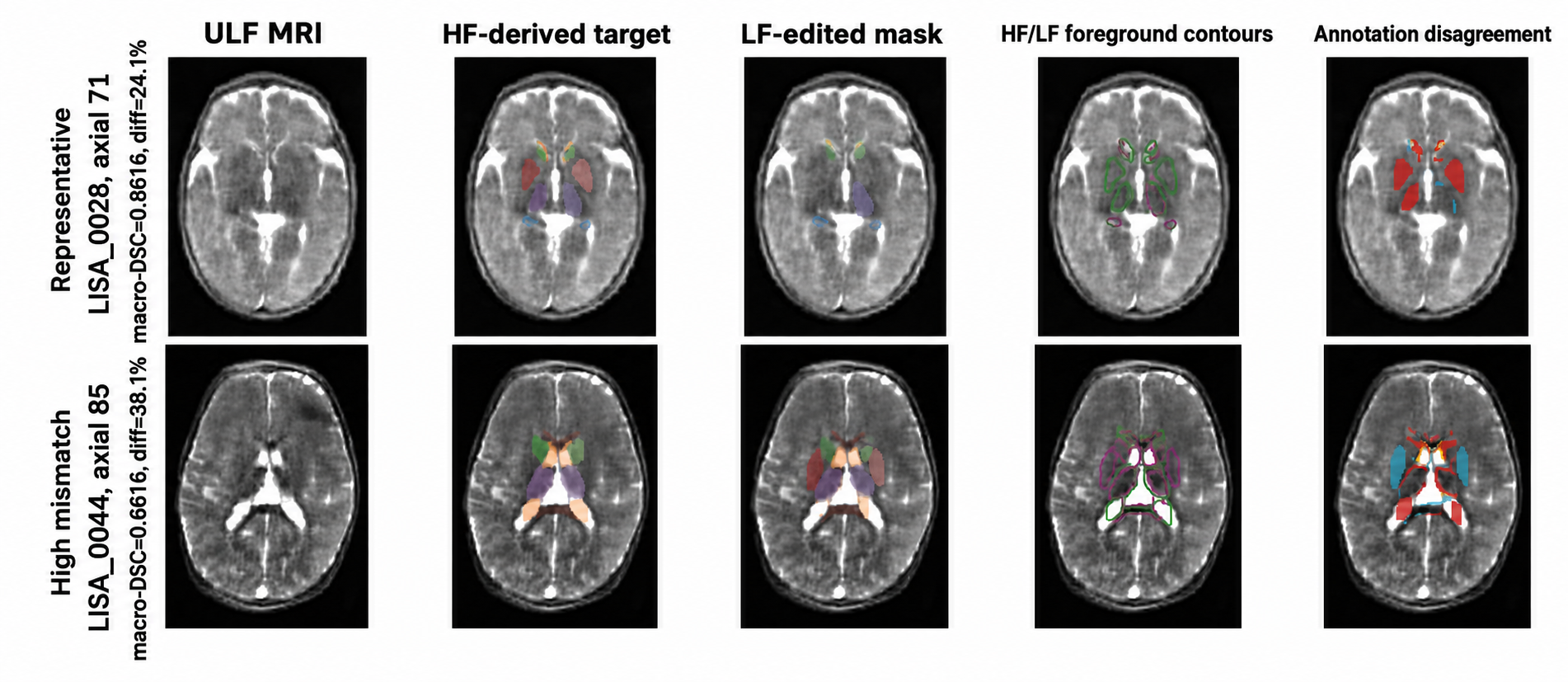}
\caption{Examples of paired LISA annotations on ULF MRI. The representative training case and the highest-mismatch training case were selected quantitatively. Green solid contours indicate the HF-derived target and magenta dashed contours the LF-edited mask. In the disagreement maps, red is HF only, cyan is LF only, and yellow indicates different non-background class assignments. Reported disagreement is the fraction of foreground-union voxels with unequal labels over the full volume.}
\label{fig:dataset}
\end{figure}

\subsection{Paired-Annotation Structure}

The two released annotations are related but not equivalent. The HF-derived mask defines the scored target but may contain local registration displacement, whereas the LF-edited mask better reflects visible ULF anatomy but may omit or simplify structures that are difficult to resolve at 0.064~T. Thus, the task is not only an image-quality problem but also a paired-supervision problem.

Figure~\ref{fig:dataset} illustrates this structure using quantitatively selected examples. LISA\_0028 is the median training case by macro HF/LF Dice over 11 labels (0.8616), while LISA\_0044 has the lowest macro-Dice (0.6616) and the largest disagreement fraction (38.1\%). For each case, the displayed axial slice contains the most disagreeing voxels. Disagreement is structured rather than random, concentrating near anatomical boundaries and sometimes affecting most of a structure on a slice. This supports treating the LF mask as useful image-aligned evidence, but not as a replacement for the HF-derived challenge target.

\subsection{Development Evaluation}

All predictions are evaluated against the Dataset001 HF masks on the same 16 development cases. We report DSC, HD, HD95, ASSD, signed RVE, and absolute RVE, covering overlap, boundary error, surface distance, and volume bias. Metrics are computed per case and foreground class, then macro-averaged. Higher DSC is better; distance metrics and absolute RVE are minimized.

\section{Methodology}

\begin{figure}[!htbp]
\centering
\includegraphics[width=\textwidth]{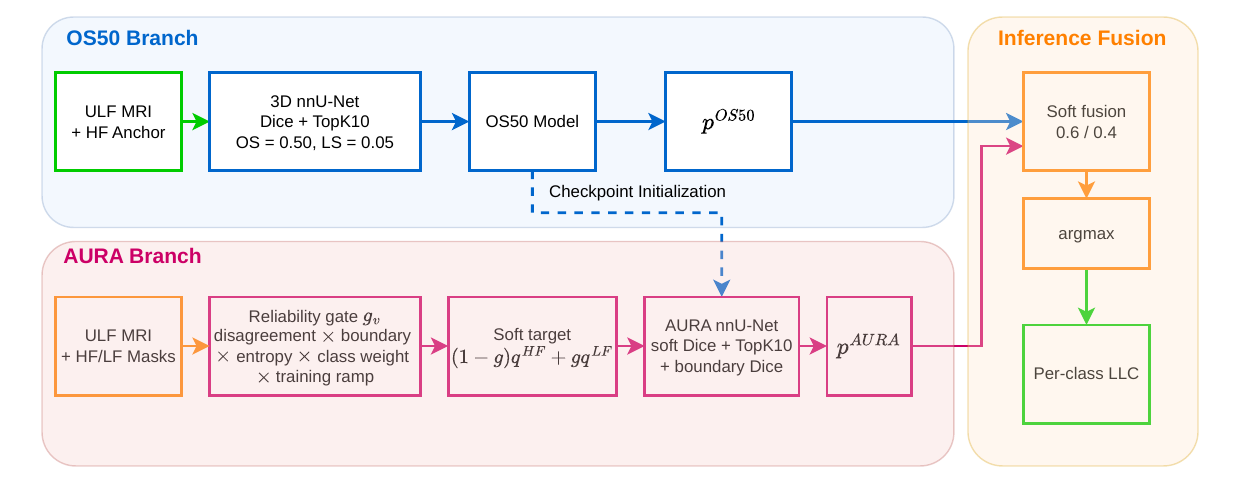}
\caption{Overview of the proposed framework. The HF-derived mask is kept as the anchor target, while the LF-edited mask contributes to AURA only through a bounded reliability gate. This gate depends on HF/LF disagreement, anatomical boundaries, predictive uncertainty, class-specific factors, and training stage. At inference, the independently trained probability maps are averaged before argmax prediction and per-class largest-connected-component filtering.}
\label{fig:framework}
\end{figure}

Figure~\ref{fig:framework} summarizes the framework. We train two nnU-Net-based predictors with the same 3D full-resolution architecture. \osfifty{} is an HF-supervised anchor trained on the HF targets. \aura{} is initialized from the best \osfifty{} checkpoint and fine-tuned on paired HF/LF annotations. The paired masks are used only to define training supervision. At inference, both predictors receive only the ULF MRI, and the predictors are combined by averaging their probability maps.

\subsection{Hard-Voxel Anchor Model}

Both predictors use the 3D full-resolution nnU-Net v2 configuration with deep supervision~\cite{isensee2021nnunet}. The anchor model, \osfifty{}, is initialized from a TS850 MRI checkpoint from TotalSegmentator~\cite{totalsegment} and trained directly against the HF-derived target. Its objective combines soft Dice loss with TopK cross-entropy computed over the hardest 10\% of valid voxels. This hard-voxel formulation focuses optimization on ambiguous regions, small structures, and boundary voxels, instead of allowing the loss to be dominated by easy background and confidently segmented interiors. The cross-entropy term uses label smoothing with $\epsilon=0.05$, while the Dice loss excludes the background class and uses a smoothing constant of $10^{-5}$. Foreground patch oversampling is increased from 0.33 to 0.50 to expose the model more frequently to small anatomical structures during training. Overall, \osfifty{} is designed as a conservative but strong anchor: it optimizes the scored HF-derived target directly, while hard-voxel learning, label smoothing, and increased foreground sampling help reduce overconfidence and improve sensitivity around difficult anatomical boundaries.

\subsection{AURA: Asymmetric Paired-Annotation Supervision}

The \aura{} predictor is initialized from the best \osfifty{} checkpoint and fine-tuned using paired HF/LF labels. Let $p_{v,c}$ be the predicted probability of class $c$ at voxel $v$. The HF label is converted to a smoothed one-hot target $q^{\mathrm{HF}}_{v,c}$ with $\epsilon=0.03$, and the LF label to a one-hot target $q^{\mathrm{LF}}_{v,c}$. The LF contribution is controlled by three detached voxel-wise reliability factors: HF/LF label disagreement $d_v$, membership in either
  label boundary $b_v$, and normalized predictive entropy $u_v$. These are defined as
  \begin{equation}
  \begin{aligned}
  d_v &= \mathbb{1}\!\left[y^{\mathrm{HF}}_v \ne y^{\mathrm{LF}}_v\right],\\
  b_v &= \mathbb{1}\!\left[v \in \partial y^{\mathrm{HF}} \cup \partial y^{\mathrm{LF}}\right],\\
  u_v &= \operatorname{stopgrad}\!\left(
  -\frac{1}{\log C}\sum_{c\in\mathcal{C}} p_{v,c}\log(p_{v,c}+\delta)
  \right).
  \end{aligned}
  \label{eq:aura-factors}
  \end{equation}
  Here $\mathcal{C}$ contains all 12 classes including background, $C=|\mathcal{C}|$, and $\partial y$ is the $3\times3\times3$ label-
  boundary mask. The epoch ramp
  $r(t)=\operatorname{clip}((t-25)/100,0,1)$ delays LF supervision until after a
  25-epoch HF-only warm-up. The final gate is
  \begin{equation}
  \begin{aligned}
  g_v = \operatorname{stopgrad}\Big(
  \operatorname{clip}\big[
  &0.35\, r(t)\, w_{y^{\mathrm{HF}}_v}
  (0.12+0.88d_v)\\
  &\times(0.35+0.65b_v)(0.35+0.65u_v),
  \,0,\,0.42\big]\Big).
  \end{aligned}
  \label{eq:gate}
  \end{equation}
  The class factor $w$ is indexed by the HF label. It sets background to zero, increases LF influence for hippocampi (1.20),
  slightly reduces it for caudate and lentiform nuclei (0.95), and further reduces it for ventricles (0.85), thalami (0.90), and
  corpus callosum (0.80). Thus, LF edits cannot create foreground where the HF anchor is background.

The gate is intentionally multiplicative. If HF and LF agree, if the voxel is far from a label boundary, or if the model is already confident, the LF term is reduced even after the warm-up. If all three signals are active, LF can make a larger contribution, but the gate remains bounded by 0.42, so the HF anchor always retains the majority of the target mass. The entropy and gate are computed without gradient flow, so the network cannot reduce its loss by artificially changing the reliability signal. In implementation, all target construction is performed inside the loss function and the nnU-Net network architecture is unchanged.

The detached gate defines a voxel-wise interpolation between the HF-derived anchor target and the LF-edited auxiliary observation:
\begin{equation}
q_v=(1-g_v)q^{\mathrm{HF}}_v+g_v q^{\mathrm{LF}}_v,
\label{eq:soft-target}
\end{equation}
so LF is a conditional auxiliary observation rather than a second ground truth. The \aura{} objective combines class-weighted soft Dice over the foreground classes, soft TopK10 cross-entropy, and a boundary-restricted soft Dice term, following work showing that boundary- and surface-aware terms complement purely region-based overlap losses~\cite{kervadec2021boundary,karimi2020hausdorff}:
\begin{equation}
\mathcal{L}_{\mathrm{AURA}} =
\mathcal{L}_{\mathrm{Dice}}(p,q)
+
\mathcal{L}_{\mathrm{TopK10}}(p,q)
+
0.03\,\mathcal{L}_{\mathrm{boundary}}(p,q).
\label{eq:aura-loss}
\end{equation}
The hippocampal Dice terms are weighted 1.25 and the corpus-callosum term 1.15 before normalization. \aura{} uses the same 0.50 foreground oversampling as the anchor model, and left--right mirroring is disabled to avoid exchanging paired anatomical labels.

Several implementation choices are deliberately conservative. AURA is not a new network, does not add a second decoder, and does not use the LF mask at inference. More generally, neither the HF mask, the LF mask, nor a high-field image is required at deployment; the only inference input is the ULF MRI. The principal change is the training target used by the loss. This makes the comparison to the HF anchor interpretable: any difference between the two predictors comes from the paired-label objective, initialization from \osfifty{}, and the disabled left--right mirroring used for paired anatomical labels. Deep supervision wraps the same loss at decoder outputs using the standard nnU-Net weights, with the lowest-resolution output assigned zero weight.

\subsection{Probability Fusion and Post-processing}

At inference, \osfifty{} and \aura{} each produce a 12-channel probability tensor. The final map is computed by weighted probability averaging:
\begin{equation}
p^{\mathrm{ens}}_v =
0.6\,p^{\mathrm{OS50}}_v
+
0.4\,p^{\mathrm{AURA}}_v,
\qquad
\hat{y}_v =
\arg\max_{c\in\mathcal{C}} p^{\mathrm{ens}}_{v,c}.
\label{eq:ensemble}
\end{equation}

After argmax, per-class largest-connected-component filtering keeps only the largest 26-connected component for each foreground label; on the development split every predicted label already forms a single component, so this step acts as a safeguard rather than an active correction. No test-time volume calibration, class-specific thresholding, or probability rescaling is applied. The 0.6/0.4 weight was selected on the development split by screening weights from 0 to 1 in increments of 0.1, so the ensemble result is validation-tuned.

\section{Experiments and Results}

\subsection{Implementation and Reproducibility}

All volumes are resampled by the nnU-Net plan to 1~mm isotropic resolution. Training uses $112\times160\times128$ voxel patches with batch size two. The full-resolution PlainConvUNet has six encoder stages with 32, 64, 128, 256, 320, and 320 feature channels; each encoder/decoder stage contains two 3$\times$3$\times$3 convolutions followed by instance normalization and leaky ReLU. Deep supervision is applied at decoder resolutions with decreasing weights, and the lowest-resolution output receives zero weight. These settings are shared by \osfifty{} and \aura{}.

% Training uses SGD for 1000 epochs with initial learning rate 0.01, polynomial decay, weight decay $3\times10^{-5}$, standard nnU-Net augmentations, and 0.50 foreground oversampling. \osfifty{} is initialized from the TotalSegmentator MRI checkpoint~\cite{totalsegment}; \aura{} starts from the best fold-0 \osfifty{} checkpoint and uses a 25-epoch HF-anchor warm-up before LF influence ramps up. At inference, nnU-Net tiled prediction with Gaussian weighting is used. \osfifty{} uses test-time mirroring over axes $(0,1,2)$, whereas \aura{} disables the left--right axis and uses axes $(0,1)$. Predictions are generated after probability fusion and checked for 12 channels, labels 0--11, preserved NIfTI geometry, and one prediction per case.

Training uses SGD with an initial learning rate of 0.01, polynomial decay, weight decay $3\times10^{-5}$, standard nnU-Net augmentations, and 0.50 foreground oversampling. \osfifty{} is trained for 1000 epochs and initialized from the TotalSegmentator MRI checkpoint~\cite{totalsegment}; \aura{} is trained for 500 epochs, initialized from the best fold-0 \osfifty{} checkpoint, and uses a 25-epoch HF-anchor warm-up before LF influence ramps up. At inference, nnU-Net tiled prediction with Gaussian weighting is used. \osfifty{} uses test-time mirroring over axes $(0,1,2)$, whereas \aura{} disables the left--right axis and uses axes $(0,1)$. Predictions are generated after probability fusion and checked for 12 channels, labels 0--11, preserved NIfTI geometry, and one prediction per case.

\subsection{Quantitative Comparison}

\begin{table}[!htbp]
\caption{Macro-averaged development results over 16 cases and all 11 foreground labels. Best values are bold. For signed RVE, the value closest to zero is bolded.}
\label{tab:main}
\centering
\resizebox{0.75\textwidth}{!}{%
\begin{tabular}{lrrrrr}
\toprule
Method & DSC $\uparrow$ & HD $\downarrow$ & HD95 $\downarrow$ & ASSD $\downarrow$ & RVE \\
\midrule
\osfifty{} & 0.7984 & 3.5245 & \textbf{1.8822} & 0.7873 & -0.0109 \\
\aura{} & 0.7950 & 3.5902 & 1.9041 & 0.7988 & \textbf{0.0036} \\
\method{} & \textbf{0.7988} & \textbf{3.5243} & 1.8892 & \textbf{0.7855} & -0.0093 \\
\bottomrule
\end{tabular}}
\end{table}

Table~\ref{tab:main} shows that the final ensemble achieves the best DSC and ASSD, with small gains over \osfifty{}, while HD95 does not improve and the ensemble does not have the best signed RVE. AURA alone is weaker than the HF-supervised baseline on the global average, but its errors appear complementary enough to yield a modest fusion gain. The absolute DSC change from \osfifty{} to the final ensemble is only $0.0004$ (0.7984 to 0.7988), and ASSD changes from 0.7873 to 0.7855, while HD95 is slightly worse. These differences are too small to support a claim of practically or clinically meaningful superiority over the HF-supervised nnU-Net baseline. Instead, we interpret them narrowly as preliminary evidence that LF-edited annotations can provide complementary image-aligned information when used as a bounded auxiliary signal rather than as a replacement for the HF-derived target. The ensemble also exceeds the matched Dice--TopK10 baseline (DSC 0.7971) evaluated under the same protocol. A label-wise ensemble reached 0.7994 DSC, but is excluded to avoid additional validation-set selection bias. Importantly, the 0.6/0.4 fusion weight and model choices were selected using the same 16-case development fold on which these results are reported. The ensemble improvement is therefore validation-tuned and should not be interpreted as an unbiased estimate of generalization.

\begin{figure}[!h]
\centering
\includegraphics[width=\textwidth]{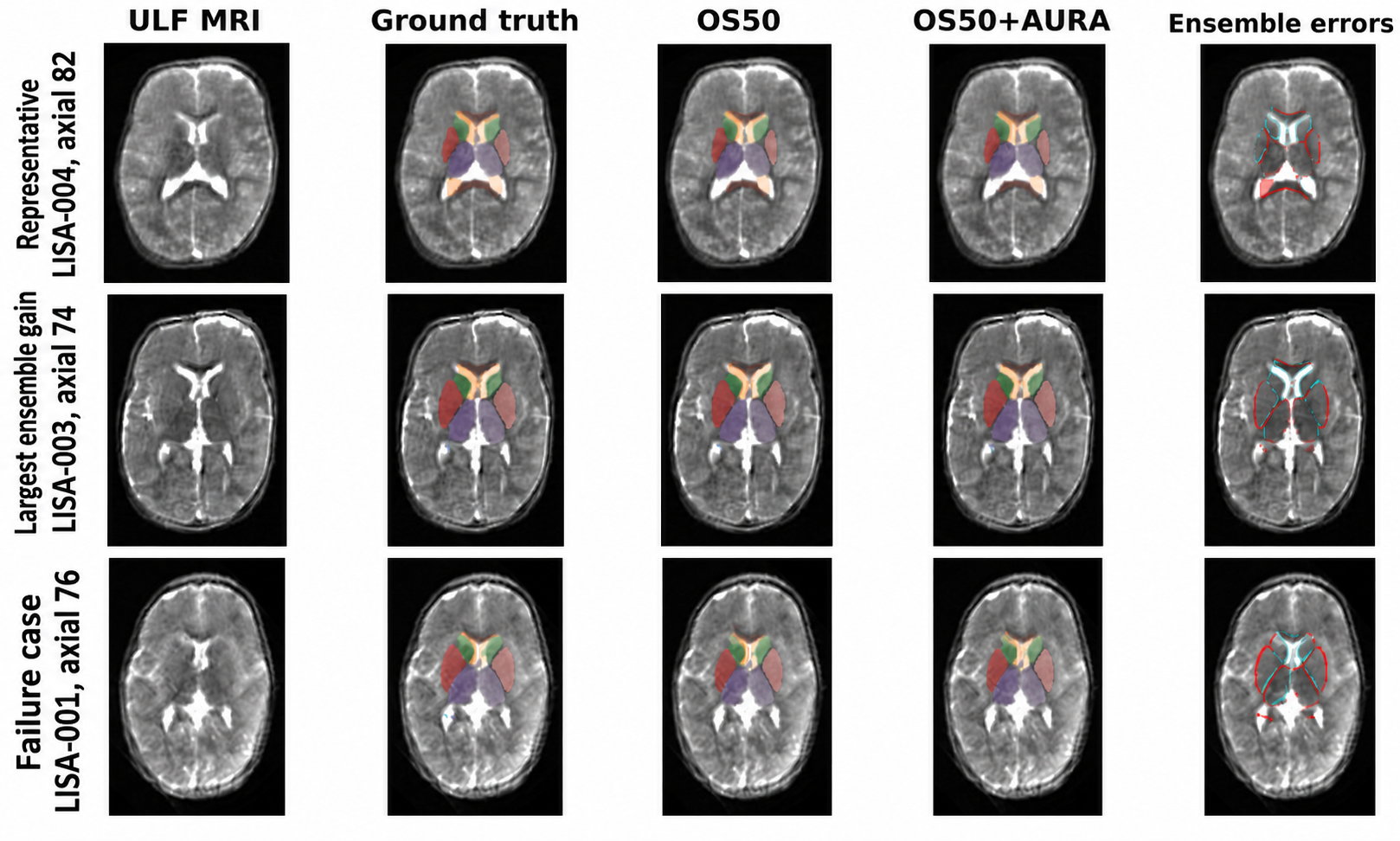}
\caption{Qualitative results. Columns show ULF MRI, reference,
OS50, ensemble, and errors. Red marks missed/misclassified reference voxels;
cyan marks false foreground. Rows show representative case, largest-gain case,
and lowest-DSC case.}
\label{fig:qual}
\end{figure}

\subsection{Qualitative Analysis}

% Figure~\ref{fig:qual} uses a pre-specified selection rule rather than visual
% cherry-picking. LISA\_0041 is near the median ensemble DSC, LISA\_0043 has the
% largest positive DSC change over OS50 ($+0.00326$), and LISA\_0021 has the
% lowest ensemble DSC. For each case, the axial slice containing the most
% ground-truth foreground voxels is displayed. The representative and improved
% cases show close agreement across large central structures, with remaining
% errors concentrated along narrow boundaries. In the failure case, weak local
% contrast produces broader errors across several bilateral structures. AURA
% does not eliminate this failure mode.

Figure~\ref{fig:qual} uses a pre-specified selection rule rather than visual cherry-picking. LISA\_0041 is near the median ensemble DSC, LISA\_0043 has the largest positive DSC change over OS50 ($+0.00326$), and LISA\_0021 has the lowest ensemble DSC. For each case, the axial slice containing the most ground-truth foreground voxels is displayed. The representative and improved cases show close agreement across large central structures, with remaining errors concentrated along narrow boundaries and small peripheral regions. In the improved case, the refinements are confined to these boundary regions, where the baseline already recovers the overall anatomy. In the failure case, weak local contrast produces broader errors across several bilateral structures, and fusion changes the outcome only marginally (DSC $+0.0001$ over \osfifty{}). AURA does not address this failure mode.

\section{Limitations}

\begin{figure}[!htbp]
\centering
\includegraphics[width=\textwidth]{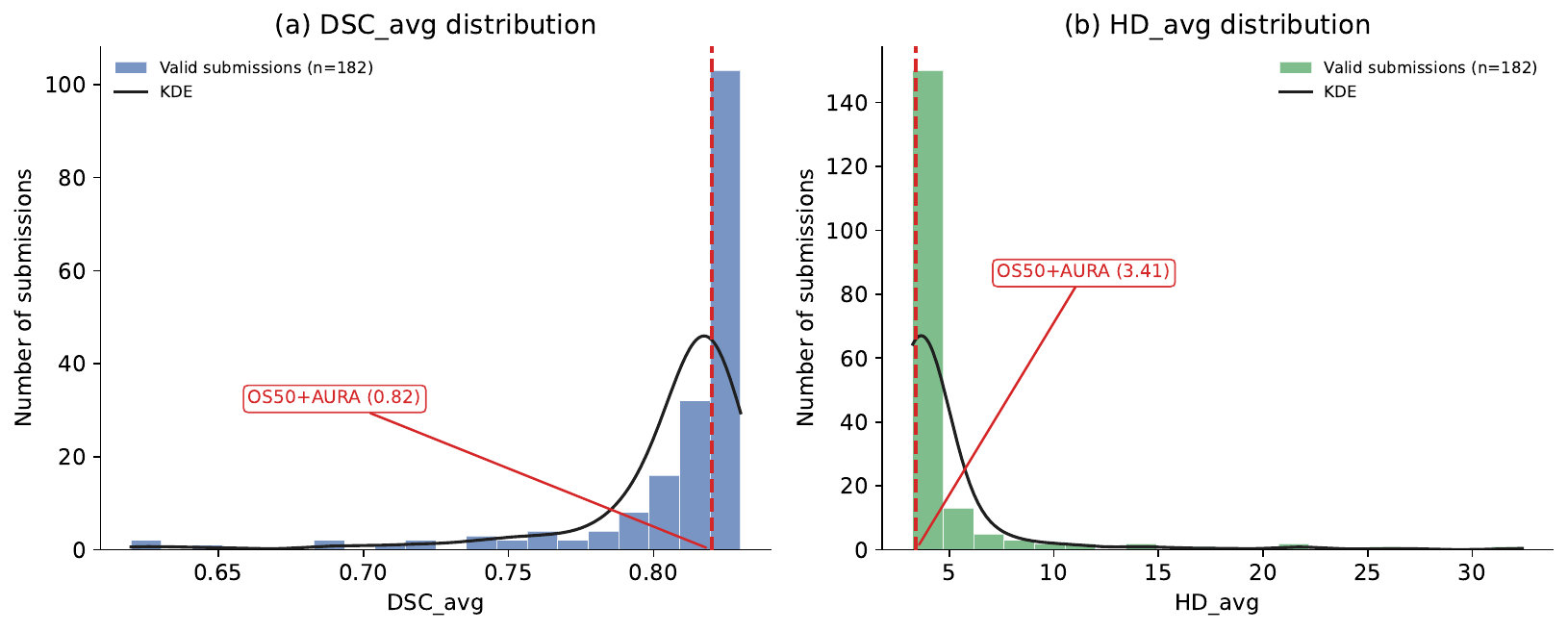}
\caption{Distribution of valid submissions on the LISA 2026 Task~2 validation
leaderboard~\protect\footnotemark, for (a)~DSC$_{\mathrm{Avg}}$ and (b)~HD$_{\mathrm{Avg}}$. The dashed line marks the \method\space submission
(DSC$_{\mathrm{Avg}}=0.82$, HD$_{\mathrm{Avg}}=3.41$), which lies near the leading end of both distributions. Source data are from the official LISA 2026 challenge leaderboard~\cite{lisa2026challenge}.}
\label{fig:val_distribution}
\end{figure}

The findings should be interpreted in the context of the LISA 2026 challenge. AURA was developed and evaluated as a challenge method rather than as a standalone study designed to establish statistically significant superiority. A key limitation is the small separation among high-performing validation submissions. As shown in Figure~\ref{fig:val_distribution}, many leading submissions fall within a narrow DSC$_{\mathrm{Avg}}$ range of approximately $0.82$--$0.83$. The \method\space submission ($0.82$ DSC$_{\mathrm{Avg}}$, $3.41$ HD$_{\mathrm{Avg}}$) lies within this region. These values are computed on hidden validation cases and are therefore not directly comparable to the 16-case development results in Table~\ref{tab:main}. Accordingly, the results support the competitiveness of AURA within the challenge setting but not strong claims of practically meaningful superiority. Moreover, the distribution is over submissions rather than distinct teams or methods, so it reflects leaderboard density rather than methodological diversity. The current evaluation assesses AURA as a whole and does not isolate the contribution of individual reliability-gate factors; factor-wise ablations are left for future work. In addition, evaluation is restricted to the challenge data, and the robustness of the method to variations in acquisition conditions, anatomy, and annotation quality remains unknown. External ULF cohorts would therefore be valuable for determining whether the observed behavior generalizes beyond the specific LISA 2026 setting. Final performance will be assessed on the hidden test set under the LISA 2026 challenge protocol.

\setcounter{footnote}{-1}

\section{Conclusion}
\footnotetext{Of the 191 submissions accepted by Synapse, 182 remained valid after excluding submissions with $\mathrm{inf}$ or $\mathrm{NaN}$ HD$_{\mathrm{Avg}}$ values.}
We presented AURA, an asymmetric dual-observation supervision strategy developed for the LISA 2026 Challenge for 11-class ULF pediatric brain MRI segmentation. AURA uses the HF-derived annotation as the primary supervision target while incorporating LF-edited annotations through a bounded reliability gate, treating the two annotations as distinct observations rather than interchangeable ground truths. The paired annotations are used only during training; at inference, the model operates solely on the ULF MRI without requiring annotations or a high-field image. Within the challenge development setting, AURA and its ensemble achieved competitive performance, although differences relative to the HF-supervised baseline were small. These results provide preliminary evidence that asymmetric use of paired annotations may offer complementary LF supervision for ULF segmentation. However, the current evaluation is limited to the available development data and does not establish superiority, robustness, or generalization beyond the challenge setting. Further evaluation on the hidden test set and external ULF cohorts, together with component-wise analysis of the reliability gate, is required to assess the broader utility of the approach.

\begin{credits}
\subsubsection{\ackname}
The computational resources used in this research were provided by the VinUni-Illinois Smart Health Center (VISHC). This research was supported by VinUniversity’s Seed Grant Program under Project VUNI.2425.EME.005.
\end{credits}

\bibliographystyle{splncs04}
\bibliography{ref}

\end{document}